\documentclass[letterpaper]{article} 
\usepackage{aaai24}  
\usepackage{times}  
\usepackage{helvet}  
\usepackage{courier}  
\usepackage[hyphens]{url}  
\usepackage{graphicx} 
\usepackage{natbib}  
\usepackage{caption} 
\usepackage{algorithm}
\usepackage{algorithmic}

\usepackage{newfloat}
\usepackage{listings}
\DeclareCaptionStyle{ruled}{labelfont=normalfont,labelsep=colon,strut=off} 
\floatstyle{ruled}
\newfloat{listing}{tb}{lst}{}
\floatname{listing}{Listing}
\title{Chinese Spelling Correction as Rephrasing Language Model}
\author{
    Linfeng Liu\equalcontrib,
    Hongqiu Wu\equalcontrib,
    Hai Zhao
}
\affiliations{
    Department of Computer Science and Engineering, Shanghai Jiao Tong University\\

    Key Laboratory of Shanghai Education Commission for Intelligent Interaction\\
    and Cognitive Engineering, Shanghai Jiao Tong University, Shanghai, China\\
    \{linfengliu, wuhongqiu\}@sjtu.edu.cn,
    zhaohai@cs.sjtu.edu.cn
    
}

\usepackage{bibentry}

\usepackage{graphicx}
\usepackage{color}
\usepackage{multirow,booktabs}
\usepackage{caption}
\usepackage{subcaption}

\begin{document}

\maketitle

\begin{abstract}
This paper studies Chinese Spelling Correction (CSC), which aims to detect and correct the potential spelling errors in a given sentence.
Current state-of-the-art methods regard CSC as a sequence tagging task and fine-tune BERT-based models on sentence pairs.
However, we note a critical flaw in the process of tagging one character to another, that the correction is excessively conditioned on the error.
This is opposite from human mindset, where individuals rephrase the complete sentence based on its semantics, rather than solely on the error patterns memorized before.
Such a counter-intuitive learning process results in the bottleneck of generalizability and transferability of machine spelling correction.
To address this, we propose \textit{Rephrasing Language Model (ReLM)}, where the model is trained to rephrase the entire sentence by infilling additional slots, instead of character-to-character tagging.
This novel training paradigm achieves the new state-of-the-art results across fine-tuned and zero-shot CSC benchmarks, outperforming previous counterparts by a large margin.
Our method also learns transferable language representation when CSC is jointly trained with other tasks.
\end{abstract}

\section{Introduction}

Chinese Spelling Correction (CSC) is a fundamental natural language processing task
to detect and correct the potential spelling errors in a given Chinese text \citep{DBLP:conf/acl-sighan/YuL14,DBLP:journals/ijclclp/XiongZZHC15}.
It is crucial for many downstream applications, e.g, named entity recognition \citep{DBLP:journals/corr/abs-2305-05253}, optical character recognition \citep{DBLP:conf/lrec/AfliQWS16}, web search \citep{DBLP:conf/tal/MartinsS04, DBLP:conf/coling/GaoLMQS10}.

Current state-of-the-art methods regard CSC as a sequence tagging task and fine-tune BERT-based models on sentence pairs \citep{DBLP:conf/naacl/DevlinCLT19}. On top of this, phonological and morphological features are further injected to enhance the tagging process \citep{DBLP:conf/acl/HuangLJZCWX20,DBLP:conf/acl/LiuYYZW20,DBLP:journals/corr/abs-2203-10929}.

\begin{figure}[t]
    \centering
    \includegraphics[width=0.46\textwidth]{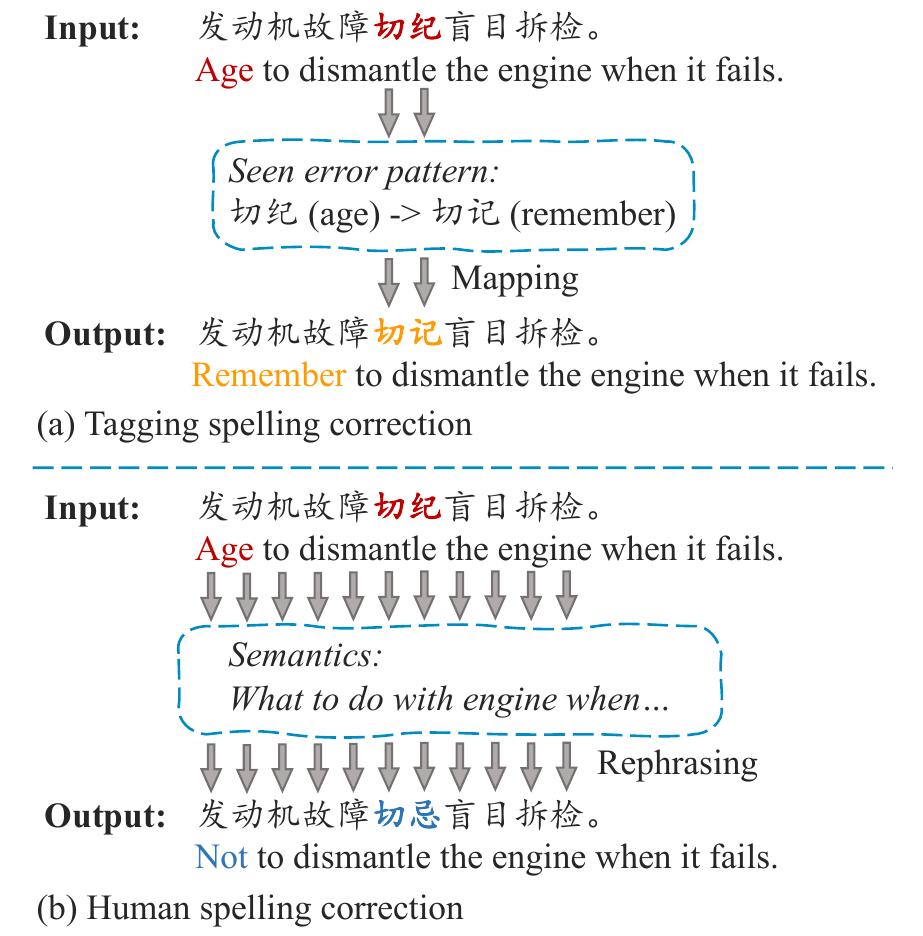}
    \caption{Comparison of tagging spelling correction and human spelling correction.}
    \label{fig:bert-human}
\end{figure}
While sequence tagging has become the prevailing paradigm in CSC,
in this paper, we note a critical flaw coming with performing character-to-character tagging, which is counter to natural human mindset.
CSC is a special form of tagging, where the majority of characters are the same between the source and target sentences.
As a result, the model is allowed to greatly memorize those mappings between the error and correct characters during training and simply copy the other characters, to still achieve a decent score on the test set.
It means that the resultant correction will be excessively conditioned on the original error itself, while ignoring semantics of the entire sentence.
We showcase a concrete example in Figure \ref{fig:bert-human}.
In (a), the model has been exposed to an edit pair (correct \textit{age} to \textit{remember}) during previous training. Now it encounters a new error \textit{age} during testing and still corrects it to \textit{remember}.
It reflects that the tagging model doggedly memorizes the training errors and fails to fit in the context.
However, this is far from human spelling correction.
In (b), when a person sees a sentence, he first commits the underlying semantics to his mind. Then, he rephrases the sentence based on his linguistic knowledge from the past, though this process will not be explicitly written down, and eventually decides how the sentence should be corrected. For example, here, it is easy for a person to correct \textit{age} to \textit{not} based on the given semantics.

Instead of character-to-character tagging, we propose to rephrase the sentence as the training objective for fine-tuning CSC models.
We denote the resultant model as \textit{Rephrasing Language Model (ReLM)}, where the source sentence will first be encoded into the semantic space, and then is rephrased to the correct sentence based on the given mask slots.
ReLM is based on BERT \citep{DBLP:conf/naacl/DevlinCLT19}, and achieves the new state-of-the-art results on existing benchmarks, outperforming previous counterparts by a large margin.
We find that the rephrasing objective also works on auto-regressive models like GPT \citep{DBLP:conf/nips/BrownMRSKDNSSAA20} and Baichuan \citep{DBLP:journals/corr/abs-2309-10305}, but worse than ReLM.

As opposed to previous work, we also pay attention to the CSC performance in multi-task settings, where CSC is jointly trained with other tasks (e.g. sentiment analysis, natural language inference).
We find that tagging-based fine-tuning leads to non-transferable language representation and the resultant CSC performance will significantly degenerate once there is another task.
The explanation still lies in the excessive condition on the errors. This problematic property makes CSC hard to be incorporated into multi-task learning.
Given the ongoing trend of instruction tuning across diverse tasks \citep{DBLP:journals/corr/abs-2303-08774}, this phenomenon has a significant negative impact.
We show that ReLM allows for better transferability between CSC and other tasks, building promising multi-task models.

Our contributions are summarized below:
(1) We propose ReLM to narrow the gap between machine spelling correction and human spelling correction.
(2) ReLM significantly enhances the generalizability of CSC models and refreshes the new state-of-the-art results across fine-tuned and zero-shot CSC benchmarks.
(3) We probe into and enhance the transferability of CSC to other tasks.
(4) Our analysis shows that ReLM effectively exploits and retains the pre-trained knowledge within PLMs, while tagging models do not. \footnote{https://github.com/Claude-Liu/ReLM}\footnote{https://github.com/gingasan/lemon}

\section{Related Work\label{related-work}}

Most of the early efforts in CSC focus on the unsupervised techniques by evaluating the perplexity of the sentence \citep{DBLP:conf/acl-sighan/YehLWCS13,DBLP:conf/acl-sighan/YuL14,DBLP:conf/acl-sighan/XieHZHHCH15}. \citet{DBLP:conf/sigir/ZhouPK19} reformulates the spell correction problem as a machine translation task.
Recent methods model CSC as a sequence tagging problem that maps each character in the sentence to the correct one \citep{DBLP:conf/emnlp/WangSLHZ18, DBLP:conf/acl/WangTZ19}. 
On top of pre-trained language models (PLMs), a number of BERT-based models with the sequence tagging training objective are proposed.
\citet{DBLP:conf/acl/ZhangHLL20} identify the potential error characters by a detection network and then leverage the soft masking strategy to enhance the eventual correction decision.
\citet{DBLP:conf/acl/ZhuYZM22} use a multi-task network to minimize the misleading impact of the misspelled characters \citep{DBLP:conf/acl/ChengXCJWWCQ20}.
There is also a line of work that incorporates phonological and morphological knowledge through data augmentation and enhances the BERT-based encoder to assist mapping the error to the correct one \citep{DBLP:conf/acl/GuoNWZX21,DBLP:conf/acl/LiZZH20,DBLP:conf/acl/LiuYYZW20,DBLP:conf/acl/ChengXCJWWCQ20,DBLP:conf/acl/HuangLJZCWX20,DBLP:conf/acl/ZhangPZWHSWW21}.
However, our method achieves the new state-of-the-art results over all these variants with the original BERT architecture, by repurposing the training objective.

Similar as previous methods, our method is based on the (PLMs) \citep{DBLP:conf/naacl/DevlinCLT19,DBLP:conf/nips/BrownMRSKDNSSAA20,DBLP:journals/corr/abs-1907-11692,DBLP:conf/emnlp/WuDZCXHZ22,DBLP:conf/iclr/HeGC23}.
However, we maximize the pre-trained power by continually optimizing the language modeling objective, instead of sentence or token classification.
Our method works on both encoder and decoder architectures, and furthermore, we discuss CSC as a sub-task in multi-task learning, which is not discussed in previous work.

More recently, \citet{DBLP:conf/acl/WuZZZ23} decompose a CSC model into two parallel models, a language model (LM) and an error model (EM), and find that tagging models lean to over-fit the error model while under-fit the language model.
An effective technique masked-fine-tuning is thus proposed to facilitate the learning of LM.
While the masking strategy is still effective in our method,
their work differs from ours in terms of bottom logic.
The masked-fine-tuned CSC model remains a tagging model, which partially mitigates the negative effect of EM.
More importantly, our method is a language model alone, instead of two parallel models. It indicates that with an effective training objective, LM can possess the functionality of EM, essentially solving the over-fitting to EM.

\section{Method}

\begin{figure*}[t]
    \centering
    \includegraphics[width=0.9\textwidth]{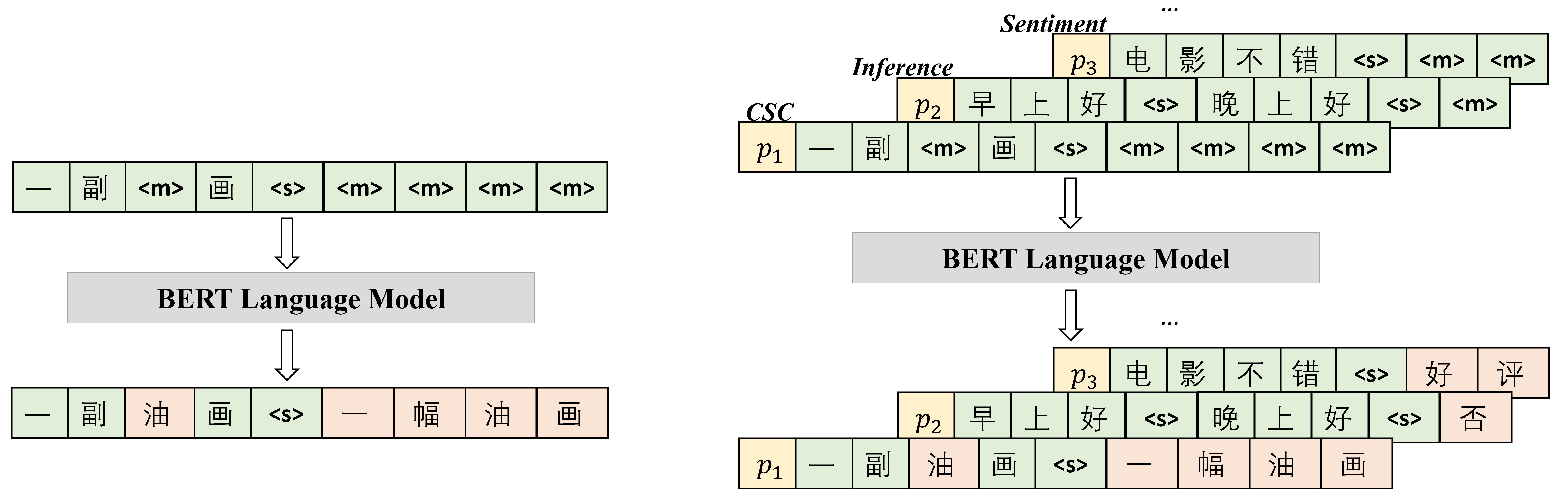}
    \caption{Paradigm of ReLM in single-task (left) and multi-task (right) settings. The source sentence for CSC is ``taking a pair ($\rightarrow$ piece) of painting'', and $\left<\rm m\right>$ and $\left<\rm s\right>$ refer to the mask and separate character respectively. On the right, we depict three tasks as a representative, CSC, language inference, and sentiment analysis, and $p$ refers to the prompt for each task.}
    \label{fig:multi-task}
\end{figure*}

\subsection{Problem Formulation}
Chinese Spelling Correction (CSC) aims to correct all misspelled characters in the source sentence.
Given a source sentence $X=\{x_1,x_2,\cdots,x_n\}$ of $n$ characters with potential spelling errors, the model seeks to generate the target sentence $Y=\{y_1,y_2,\cdots,y_n\}$ of the same length with all potential errors corrected.
The above process can be formulated as a conditional probability $P(Y|X)$.
Specifically for $x_i$, suppose that it is an error character and its ground truth is $y_i$, then the probability to correct $x_i$ to $y_i$ can be written as $P(y_i|X)$.

\subsection{Tagging}
Sequence tagging is a common model in many natural language processing tasks, where the model is trained to map one character to another correspondingly, e.g. named entity recognition, part-of-speech tagging.
All these tasks share an pivotal property in that they strongly rely on the alignment information between input and output characters.
However, deep neural models like Transformer \citep{DBLP:conf/nips/VaswaniSPUJGKP17} are always good at exploiting spurious clues if it is possible to achieve lower training loss, especially when the training data is not big enough \citep{DBLP:conf/aaai/WuDZXHZ23,DBLP:conf/emnlp/Wu0ZZ23}.
For example, \textit{Norway} is always a geopolitical entity (GPE) in entity recognition. Consequently, the model can memorize such a character-to-character mapping and make correct predictions in most situations.
Similar in CSC, the model can greatly memorize the trivial edit pair of correcting $x_i$ to $y_i$ and continue to apply it in a different context of $x_i$, without referring to the semantics.
Hence, the original training objective degenerates to:
\begin{equation}
    P(y_i|X) \approx P(y_i|x_i)
    \label{eq:tag}
\end{equation}
where we suppose $x_i$ is an error character.

However, the errors in CSC are much more diverse. As previously shown in Figure \ref{fig:bert-human}, \textit{age to} can be both corrected to \textit{remember to} or \textit{not to}, which significantly relies on the immediate context.
The resultant tagging model can hardly be generalized to unseen errors.

\subsection{Rephrasing}
In this paper, we propose to substitute sequence tagging with rephrasing as the primary training objective for CSC.

Our intuition is to eliminate the trend that the model fits the training data by naively memorizing the errors.
To do this, we train the pre-trained language models (PLMs) to rephrase the source sentence following it.
Concretely, the Transformer layers first transfer the source sentence to the semantic space. Then, the model generates a new sentence while correcting all errors in it, based on the semantics.
This process is consistent with the human process of doing spelling correction.
When a person sees a sentence, he will first commit the sentence to his mind (akin to encoding it to the semantic space), and then transform the semantics to a new sentence based on his linguistic instinct (the pre-trained weights in the PLM).
We see that the pre-training knowledge offers a great foundation for such learning rephrasing, which is aside from learning sequence tagging.
Our following experiments show that sequence tagging does not make good use of the benefits from pre-training.

The process of rephrasing can be modeled based on the auto-regressive architecture with a decoder to generate the output characters one by one, e.g. GPT \citep{DBLP:conf/nips/BrownMRSKDNSSAA20}.
Specifically, we concatenate the source characters $X$ and the target characters $Y$ as the input sentence, i.e. $\{x_1,x_2,\cdots,x_n,\left<\rm s\right>,y_1,y_2,\cdots,y_n,\left<\rm eos\right>\}$, where $\left<\rm s\right>$ and $\left<\rm eos\right>$ refers to the separate token and wrap token, and train the model to predict all the target characters $y_i$ auto-regressively.
Hence, rephrasing-based spelling correction seeks to solve the following probability for $y_i,i>=1$:
\begin{equation}
    P(y_i|X) \approx P(y_i|X,y_1,y_2,\cdots,y_{i-1}).
    \label{eq:gpt}
\end{equation}

\subsection{Rephrasing Language Model}

Based on the BERT-based architecture, we propose \textit{Rephrasing Language Model} (\textit{ReLM}), a non-auto-regressive rephrasing model.

Eq \ref{eq:gpt} generates the sentences free of lengths. Given that the length of the target sentence is fixed in CSC, equal to that of the source sentence, such freedom may bring a negative impact.
BERT is an encoder-only architecture \citep{DBLP:conf/naacl/DevlinCLT19}, pre-trained by randomly setting a portion of characters to a mask symbol $\left<\rm m\right>$.
In contrast to the auto-regressive model, which keeps generation until $\left<\rm eos\right>$, BERT is programmed to only infill the pre-set slots of mask.

As shown in Figure \ref{fig:multi-task}, we concatenate the source characters $X$ and a sequence of mask characters $M=\{m_1,m_2,\cdots,m_n\}$ of the same length as the input sentence, i.e. $\{x_1,x_2,\cdots,x_n,\left<\rm s\right>,m_1,m_2,\cdots,m_n\}$, where $m_i$ refers to the mask character for $y_i$, and train the model to infill all the mask characters $m_i$ following $X$.
Since BERT can see both the left-side and right-side context, ReLM seeks to solve the following probability for $y_i,i=1\sim n$:
\begin{equation}
    P(y_i|X) \approx P(y_i|X,m_1,m_2,\cdots,m_n).
    \label{eq:bert}
\end{equation}

ReLM is superior to the auto-regressive model as it can always generate the output sentence of the same length as the input, which makes it more accurate.
In our following experiments,
we find that both auto-regressive rephrasing and ReLM outweigh previous tagging models, and the latter achieves more powerful results.

\paragraph{Auxiliary Masked Language Modeling}
As opposed to tagging models, fine-tuned ReLM on CSC is still a language model as its core.
However, there remains a chance that the model can learn the alignment of source and target sentences.
We thus propose a key strategy, that is to uniformly mask a fraction of the non-error characters in the source sentence with an unused token, to greatly regularize the model against learning character-to-character alignment.
ReLM eventually necessitates correcting potential typos while simultaneously restoring the entire sentence.

\paragraph{Distinguish from Sequence Tagging}
ReLM is a biased estimation of $P(y_i|X)$, which optimizes $P(y_i|X,m_1,m_2,\cdots,m_n)$ instead. The resultant model is forced to rely on the entire semantics.
The key property is,
Eq. \ref{eq:bert} predicts $y_i$ conditioned on the entire source sentence $X$, in contrast to Eq. \ref{eq:tag}.
More concretely, there is no alignment of characters in ReLM, and the model is not allowed to find a shortcut to perform character-to-character mapping as it does in sequence tagging.

\subsection{ReLM for Multi-Task}

The emerging large language models \citep{DBLP:journals/corr/abs-2303-08774,DBLP:journals/corr/abs-2302-13971} tend to handle diverse tasks at the same time, and it is time to study the incorporation of CSC into other tasks.
In typical multi-task learning, we add a specific classification head for each task and train a shared encoder for all tasks.
For instance, CSC will share the same language representation within the encoder with sentence classification.
However, our empirical analysis shows that the performance of conventional tagging-based CSC may largely deteriorate when it is jointly trained with other tasks.
The corresponding probing is in the following analysis section.

In contrast, ReLM, still a language model as its core, naturally suits the multi-task learning on top of language modeling, while tagging-based CSC does not.
Concretely, each individual task is templated to the format of masked language modeling, as shown in Figure \ref{fig:multi-task}.
In general, all tasks are unified to a rephrasing-like format, which enhances the transferablity of CSC to various tasks.
In addition, ReLM supports prompt tuning \citep{DBLP:conf/emnlp/LesterAC21,DBLP:journals/corr/abs-2103-10385}. We prefix a sequence of trainable characters to the input sentence as the prompt steering the model for different tasks, and optimize the corresponding prompt for each task.
We find that introducing prompts can further improve the outcome, but to a slight extent.

\section{Experiment}

In this section, we compare ReLM with a line of tagging-based methods on existing benchmarks.
We also evaluate the CSC performance in multi-task learning, where all the models are jointly trained on three different tasks, CSC, semantic similarity, and news classification.

\subsection{Dataset}
\noindent \textbf{ECSpell} \quad
ECSpell \citep{DBLP:journals/corr/abs-2203-10929} is a CSC benchmark with three domains, LAW (1,960 training and 500 test samples), MED (3,000 training and 500 test samples), and ODW (1,728 training and 500 test samples).

\noindent \textbf{LEMON} \quad
Large-scale multi-domain dataset with natural spelling errors (LEMON) \citep{DBLP:conf/acl/WuZZZ23} is a novel CSC benchmarks with diverse real-life spelling errors. It spans 7 different domains with totally 22,252 test samples. It typically measures the open-domain generalizability of a CSC model in a zero-shot setting.

\noindent \textbf{SIGHAN} \quad
SIGHAN \citep{DBLP:conf/acl-sighan/TsengLCC15} is a CSC benchmark collected from the Chinese essays written by foreign speakers.
Following \citet{DBLP:conf/acl/WuZZZ23}, we evaluate the model on SIGHAN as zero-shot learning.

\noindent \textbf{AFQMC} \quad
Ant Financial Question Matching (AFQMC) \citep{DBLP:conf/coling/XuHZLCLXSYYTDLS20} is a Chinese semantic similarity dataset that requires the model to predict whether the given two questions are semantically similar. It contains 34,334 training samples and 3,861 test samples.

\noindent \textbf{TNEWS} \quad
TouTiao Text Classification for News Titles (TNEWS) \citep{DBLP:conf/coling/XuHZLCLXSYYTDLS20} is a text classification dataset, requiring to link each given title to 15 news categories. It contains 53,360 training samples and 10,000 test samples.

\begin{table}[t]
    \centering
    \begin{tabular}{@{}l|l|lll@{}}
    \toprule

    & Method & Prec. & Rec. & F1 \\
    \hline\hline
    \multirow{8}{*}{LAW} & GPT2\textsubscript{Tagging}         & 37.7 & 32.5 & 35.0 \\
                         & BERT\textsubscript{Tagging}         & 43.3 & 36.9 & 39.8 \\
                         & GPT2\textsubscript{Rephrasing}      & \textbf{61.6} & \textbf{84.3} & \textbf{71.2}$_{\uparrow 31.4}$ \\
                         & BERT\textsubscript{Tagging}-MFT     & 73.2 & 79.2 & 76.1 \\
                         & MDCSpell\textsubscript{Tagging}-MFT & 77.5 & 83.9 & 80.6 \\
                         & \textsc{ReLM}                       & \textbf{89.9} & \textbf{94.5} & \textbf{91.2}$_{\uparrow 10.6}$ \\
                         & Baichuan2\textsubscript{Rephrasing} & 85.1 & 87.1 & 86.0 \\
                         & ChatGPT-10 shot                     & 46.7 & 50.1 & 48.3 \\
                    
    \hline\hline
    \multirow{8}{*}{MED} & GPT2\textsubscript{Tagging}         & 23.1 & 16.7 & 19.4 \\
                         & BERT\textsubscript{Tagging}         & 25.3 & 20.0 & 22.3 \\
                         & GPT2\textsubscript{Rephrasing}      & \textbf{29.6} & \textbf{44.7} & \textbf{35.6}$_{\uparrow 13.3}$ \\
                         & BERT\textsubscript{Tagging}-MFT     & 57.9 & 58.1 & 58.0 \\
                         & MDCSpell\textsubscript{Tagging}-MFT & 69.9 & 69.3 & 69.6 \\
                         & \textsc{ReLM}                       & \textbf{79.2} & \textbf{85.9} & \textbf{82.4}$_{\uparrow 12.8}$ \\
                         & Baichuan2\textsubscript{Rephrasing} & 72.6 & 73.9 & 73.2 \\
                         & ChatGPT-10 shot                     & 21.9 & 31.9 & 26.0 \\
                         
    \hline\hline
    \multirow{8}{*}{ODW} & GPT2\textsubscript{Tagging}         & 26.8 & 19.8 & 22.8 \\
                         & BERT\textsubscript{Tagging}         & 30.1 & 21.3 & 25.0 \\
                         & GPT2\textsubscript{Rephrasing}      & \textbf{46.2} & \textbf{64.3} & \textbf{53.8}$_{\uparrow 28.8}$ \\
                         & BERT\textsubscript{Tagging}-MFT     & 59.7 & 58.8 & 59.2 \\
                         & MDCSpell\textsubscript{Tagging}-MFT & 65.7 & 68.2 & 66.9 \\
                         & \textsc{ReLM}                       & 82.4 & \textbf{84.8} & \textbf{83.6}$_{\uparrow 16.7}$ \\
                         & Baichuan2\textsubscript{Rephrasing} & \textbf{86.1} & 79.3 & 82.6 \\
                         & ChatGPT-10 shot                     & 56.5 & 57.1 & 56.8 \\
                
    \bottomrule
    \end{tabular}
    \caption{Precison, recall, and F1 results on ECSpell. We mark the performance improvement of GPT2-rephrasing over BERT-tagging and ReLM over previous SotA.}
    \label{single-ecspell}
\end{table}

\subsection{Methods to Compare}

\begin{table*}[t!]
\centering
\setlength\tabcolsep{5pt}
\begin{tabular}{llllllllll}
                       & \textbf{GAM}          & \textbf{ENC}          & \textbf{COT}          & \textbf{MEC}          & \textbf{CAR}          & \textbf{NOV}          & \textbf{NEW}          & \textbf{SIG}          & Avg           \\ \toprule[1.5pt]
\multicolumn{10}{l}{\textit{Previous tagging models SotA} \citep{DBLP:conf/acl/WuZZZ23}} \\
BERT            & 27.1                  & 41.6                  & 63.9                  & 47.9                  & 47.6                  & 34.2                  & 50.7                  & 50.6                  & 45.5          \\
BERT-MFT        & \textbf{33.3} & 45.5 & 64.1 & 50.9 & 52.3 & 36.0 & 56.0 & 53.4 & 48.9 \\
Soft-Masked-MFT & 29.8 & 44.6 & 65.0 & 49.3 & 52.0 & 37.8 & 55.8 & 53.4 & 48.4 \\
MDCSpell-MFT    & 31.2 & 45.9 & 65.4 & 52.0 & 52.6 & \textbf{38.6} & 57.3 & 54.7 & 49.7 \\
CRASpell-MFT    & 30.7 & 48.1 & 66.0 & 51.7 & 51.7 & \textbf{38.6} & 55.9 & 55.1 & 49.7 \\ \midrule
\textsc{ReLM} (\textit{Ours})   & 33.0 & \textbf{49.2}$_{\uparrow 3.7}$ & \textbf{66.8}$_{\uparrow 2.7}$ & \textbf{54.0}$_{\uparrow 3.1}$ & \textbf{53.1}$_{\uparrow 0.8}$ & 37.8$_{\uparrow 1.8}$ & \textbf{58.5}$_{\uparrow 2.5}$ & \textbf{57.0}$_{\uparrow 3.6}$ & \textbf{51.2}$_{\uparrow 2.3}$ \\ \bottomrule
\end{tabular}
\caption{Performances (F1) of ReLM and previous SotA tagging models on LEMON, where SIG refers to SIGHAN. We mark the performance improvement of ReLM over BERT-MFT.}
\label{lemon}
\end{table*}

\noindent \textbf{BERT}\textsubscript{Tagging} \quad We fine-tune the original BERT model as sequence tagging\footnote{\url{https://huggingface.co/bert-base-chinese}}.

\noindent \textbf{MDCSpell}\textsubscript{Tagging} \quad It is an enhanced BERT-based model with a detector-corrector design \citep{DBLP:conf/acl/ZhuYZM22}.




\noindent \textbf{GPT2}\textsubscript{Tagging} \quad We initialize a new classifier following the pre-trained Chinese GPT2 model and fine-tune it as sequence tagging\footnote{\url{https://huggingface.co/uer/gpt2-chinese-cluecorpussmall}}.

\noindent \textbf{Masked-Fine-Tuning (MFT)} \quad It is a simple and effective fine-tuning technique when fine-tuning tagging models, which achieves the previous state-of-the-art (SotA) results on ECSpell and LEMON \citep{DBLP:conf/acl/WuZZZ23}.

\noindent \textbf{Baichuan2-7b} \quad We fine-tune Baichuan2 \citep{DBLP:journals/corr/abs-2309-10305}, one of the strongest Chinese open source LLMs, with LoRA \citep{DBLP:conf/iclr/HuSWALWWC22}.

\noindent \textbf{ChatGPT} \quad We instruct ChatGPT \citep{DBLP:journals/corr/abs-2303-08774} to correct samples by in-context learning with 10 shots, using the \textit{openai} API\footnote{\url{gpt-3.5-turbo}}.

\noindent \textbf{\textsc{ReLM}} \quad We train ReLM based on the same BERT model in BERT-tagging.

\subsection{Fine-tuned CSC on ECSpell}

We fine-tune each model separately on the three domains for 5000 steps, with the batch size selected from \{32, 128\} and learning rate from \{2e-5, 5e-5\}.

Table \ref{single-ecspell} summarizes the results on ECSpell. We find that naive tagging models of BERT and GPT2 perform poorly on all of three domains, while BERT performs slightly better than GPT2.
However, ReLM yields amazing performance improvement. Concretely, it achieves the new SotA result on every domain, significantly outperforming the previous SotA masked-fine-tuned MDCSpell by 10.6, 12.8, and 16.7 absolute points respectively.
We also apply the rephrasing objective to GPT2. We find that even GPT2-rephrasing outperforms BERT-tagging by a large margin (e.g. F1 39.8 $\rightarrow$ 71.2 on LAW), demonstrating the great superiority of the rephrasing objective, which prevents the model from simply memorizing errors.

However, we see ReLM is more powerful.
It indicates that fixed-length rephrasing is naturally matched with CSC, while the auto-regressive one is also promising for future study.
On the other hand, it is worth noting that ReLM surpasses all other enhanced architectures by simply training based on the original Transformer architecture.
It highlights the pivotal role of the rephrasing objective, while the common tagging objective does not exploit the full power of PLMs, incurring the performance bottleneck.

We report the results of Baichuan2 and ChatGPT as representatives for LLMs.
We find that ChatGPT does not work well on CSC even in a 10-shot setting. We speculate that this is due to the lack of high-quality annotated data for CSC on the web.
However, fine-tuned Baichuan2 achieves promising results, outperforming GPT2-rephrasing by a large margin.


\subsection{Zero-Shot CSC on LEMON}

On LEMON, we evaluate models as a zero-shot learner. Following \citet{DBLP:conf/acl/WuZZZ23}, we collect 34 million monolingual sentences and synthesize training sentence pairs using the confusion set.
We train the model with the batch size 4096 and learning rate 5e-5 on 8 A800 sheets for 60,000 steps.

Table \ref{lemon} compares the zero-shot performance of ReLM to previous SotA tagging models.
We find that, though each LEMON domain varies greatly, ReLM brings a significant performance boost in almost every domain, and reaches the new SotA results, raising the previous SotA from 49.7 to 51.2.
It indicates that ReLM is more generalizable to out-of-distribution errors over all other BERT-tagging variants.

\subsection{CSC in Multi-Task}
\begin{table*}[t!]
    \centering
    \small
    \begin{tabular}{@{}l|l|ccccccc@{}}
    \toprule
        
         & \multirow{2}{*}{CSC Method} & \multicolumn{2}{c}{\small \textit{CSC}} & \multicolumn{2}{c}{\small \textit{News Classification}}  & \multicolumn{2}{c}{\small \textit{Semantic Similarity}} & \multirow{2}{*}{Avg} \\
         \cmidrule(lr){3-4}  \cmidrule(lr){5-6} \cmidrule(lr){7-8}
         & & F1 & $\Delta$ F1\textsubscript{single} (\%) & F1 & $\Delta$ F1\textsubscript{single} & F1 & $\Delta$ F1\textsubscript{single} & \\
          
    \hline\hline
    \multirow{4}{*}{LAW} & BERT\textsubscript{Tagging} & 34.5          & - 13.3\% & 56.2 & - 0.4 & 72.6 & \textbf{- 1.3} & 54.4\\
                    & BERT\textsubscript{Tagging}-MFT  & 62.0          & - 18.9\% & 55.0 & - 1.6 & 71.0 & - 2.9 &62.6\\
                    & \textsc{ReLM}                             & 84.2  & - 8.7\% & 56.9 & + 0.3 & 71.6 & - 2.3 &70.9\\
                    & \textsc{ReLM} (prompt)           & \textbf{87.6} & - \textbf{4.9\%} & 56.9 & \textbf{+ 0.3} & 72.4 & - 1.5 &\textbf{72.3} \\
    \hline\hline
    \multirow{4}{*}{MED} & BERT\textsubscript{Tagging} & 15.1          & - 32.2\% & 56.3 & - 0.3 & 72.5 & - 1.4 &48.0\\
                    & BERT\textsubscript{Tagging}-MFT  & 48.8          & - 15.9\% & 55.8 & - 0.8 & 72.5 & \textbf{- 1.3} &59.0\\
                    & \textsc{ReLM}                             & 76.1  & - 10.9\% & 57.1 & \textbf{+ 0.5} & 70.7 & - 3.2 &68.0\\
                    & \textsc{ReLM} (prompt)           & \textbf{80.8} & - \textbf{4.7\%} & 56.6 & + 0 & 71.8 & - 2.1 &\textbf{69.7}\\
    \hline\hline
    \multirow{4}{*}{ODW} & BERT\textsubscript{Tagging} & 16.8          & - 32.2\% & 56.6 & + 0   & 73.3 & \textbf{- 0.6} &48.9\\
                    & BERT\textsubscript{Tagging}-MFT  & 52.4          & - 11.5\% & 55.9 & - 0.7 & 73.2 & - 0.7 &60.5\\
                    & \textsc{ReLM}                            & 75.0  & - 13.5\% & 56.9 & \textbf{+ 0.3} & 71.8 & - 2.1 &67.9\\
                    & \textsc{ReLM} (prompt)           & \textbf{78.0} & - \textbf{10.0\%} & 56.8 & + 0.2 & 72.5 & - 1.4 &\textbf{69.1}\\ \bottomrule
    \end{tabular}
    \caption{Results of different CSC training methods in the multi-task setting (ECSpell, TNEWS, AFQMC from left to right), where $\Delta$ F1\textsubscript{single} refers to the performance difference form multi-task to single-task, and \% means the relative difference.}
    \label{multi}
\end{table*}

We train the multi-task model on three distinct tasks, ECSpell for CSC, AFQMC for semantic similarity, and TNEWS for news classification.
The three datasets are mixed together and we uniformly sample one batch from them during training.
We fine-tune each model on all tasks for 15 epochs, with the batch size selected from \{32, 128\} and learning rate from \{2e-5, 5e-5\}.
For the tagging models, we train three task-specific linear classifiers and one shared encoder.
For ReLM, we share the entirety of model parameters for all three tasks.
For ReLM with prompt embeddings, we train an additional prompt embeddings for each task.
The MFT technique is exclusively applied on CSC.

Table \ref{multi} compares the results on multiple tasks.
We find that the performances of two text classification tasks vary only marginally between multi-task and single-task settings.
We speculate that these two tasks are less challenging, and the model can fit them well more easily.
In contrast, the performance of CSC is strongly affected by other tasks, where both tagging models meet a great performance drop.
However, ReLM can largely maintain the CSC performance and achieve competitive results on all three domains, almost without compromising other tasks.
Adding additional prompt characters can further improve the performance.
It suggests that ReLM contributes to better collaboration between different tasks, on top of templating all tasks to the MLM format, while tagging-based CSC is incompatible to such a training paradigm.
The logic behind is that ReLM retains the useful features within the pre-trained language representation of PLMs.
In our following analysis, we show that tagging-based CSC learns non-transferable features.

\section{Further Analysis}

\subsection{False Positive Rate}

\begin{table}[t]
    \centering
    \small
    \begin{tabular}{@{}l|cccl@{}}
    \toprule
    Method & LAW & MED & ODW & Avg \\ \midrule
    BERT\textsubscript{Tag}                  & 13.1 & 9.1  & 13.9 & 12.0 \\
    BERT\textsubscript{Tag} (multi-task)     & 13.8 & 10.2 & 15.5 & 13.2$_\uparrow$ \\
    BERT\textsubscript{Tag}-MFT              & 14.7 & 11.2 & 15.5 & 13.8 \\
    BERT\textsubscript{Tag}-MFT (multi-task) & 14.7 & 11.6 & 18.5 & 14.9$_\uparrow$ \\
    MDCSpell\textsubscript{Tag}-MFT          & 14.3 & 10.5 & 16.4 & 13.7 \\
    \textsc{ReLM}                            & \textbf{8.4}  & \textbf{5.0}  & \textbf{6.9}  & \textbf{6.8} \\
    \textsc{ReLM} (multi-task)               & \textbf{7.4}  & \textbf{6.5}  & \textbf{2.2}  & \textbf{5.4}$_\downarrow$ \\ \bottomrule
    \end{tabular}
    \caption{Comparison of false positive rate (FPR) on ECSpell. It is excepted to be lower for a better CSC system.}
    \label{ecspell-fpr}
\end{table}

False positive rate (FPR) is a measurement to evaluate a CSC system in real-world applications, which refers to the ratio that the model mistakenly modifies an otherwise correct sentence, which is also known as over-correction.
Table \ref{ecspell-fpr} shows that ReLM greatly reduces the FPR compared to tagging models.
It suggests the tagging models are overly conditioned on the seen errors and thus tend to modify some new expressions to familiar ones, while ReLM does not.
Additionally, we find that ReLM produces even lower FPR in multi-task learning.
It indicates that by ReLM, the language representation learned from CSC and other tasks can complement each other, while sequence tagging cannot.

We further demonstrate that a high FPR may result in a gap between the development performance and real-world practice.
Mathematically, we have $1/{\rm p} \propto \frac{N}{P}\cdot{\rm FPR}$,
where $N$ and $P$ refer to the number of negative samples and positive samples, 
and ${\rm p}$ is the precision score.
We can find that $p$ is negatively correlated with the ratio $\frac{N}{P}$, which means more negative samples lead to lower precision under the same FPR.
However, negative samples are dominant in real-world situations ($\frac{N}{P}$ is large), since humans do not misspell very frequently.
We can derive similarly results for the F1 score.
Consequently, a higher FPR may exacerbate the decrease of the overall performance of the CSC system.

\begin{figure}[t]
    \centering
    \begin{subfigure}{0.23\textwidth}
        \centering
        \includegraphics[width=\textwidth]{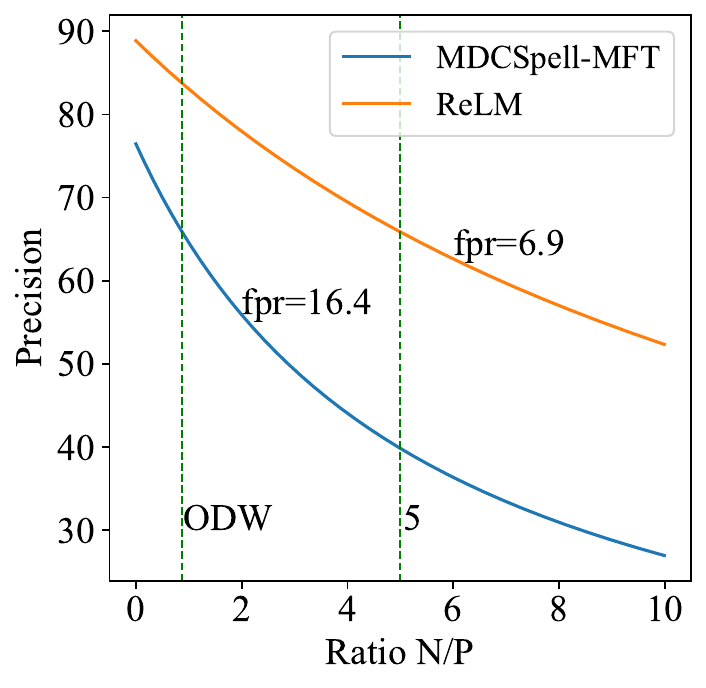}
        \caption{Precision}
        \label{fig:precision}
    \end{subfigure}
    \begin{subfigure}{0.23\textwidth}
        \centering
        \includegraphics[width=\textwidth]{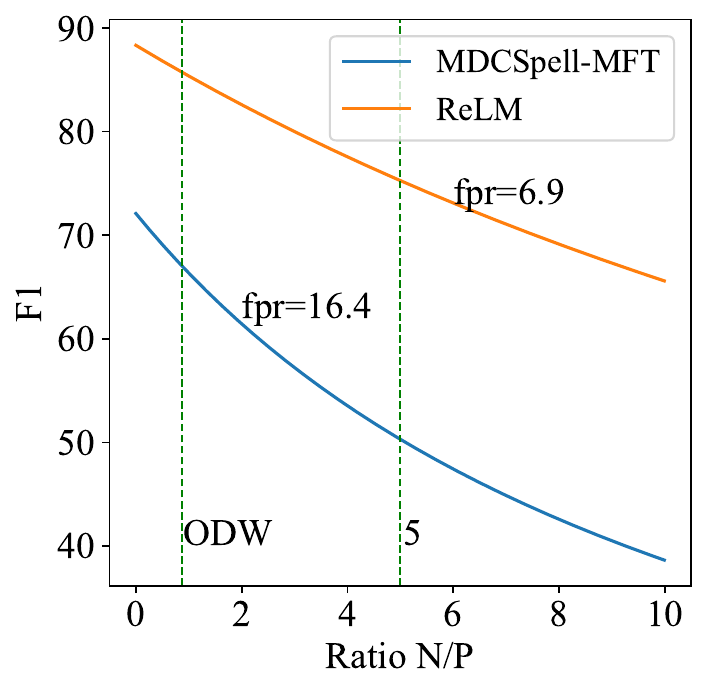}
        \caption{F1}
        \label{fig:f1}
    \end{subfigure}
    \caption{Performance variation (precision and F1) with the proportion of negative and positive samples.}
    \label{fig:fpr}
\end{figure}

Figure \ref{fig:fpr} depicts the variation of the precision and F1 with the proportion of positive and negative samples, comparing MDCSpell-MFT to ReLM on ECSpell-ODW.
We find that both F1 and precision curves of ReLM are more gentle, which decrease more slowly with the increase of $\frac{N}{P}$.
It highlights the practical value of ReLM for real applications.

\begin{table}[t]
    \centering
    \small
    \begin{tabular}{@{}l|lllc@{}}
    \toprule
    \multirow{2}{*}{CSC Method} & LAW{\rm $\rightarrow$}   & MED{\rm $\rightarrow$}  & ODW{\rm $\rightarrow$}    & \multirow{2}{*}{TNEWS} \\ & TNEWS & TNEWS & TNEWS & \\ \midrule
    BERT\textsubscript{Tag}     & 13.2$_{\downarrow 43.4}$ & 14.8$_{\downarrow 41.8}$ & 15.7$_{\downarrow 40.9}$ & 56.6 \\
    BERT\textsubscript{Tag}-MFT & 16.1$_{\downarrow 40.5}$ & 17.6$_{\downarrow 39.0}$ & 18.5$_{\downarrow 38.1}$ & 56.6 \\
    \textsc{ReLM}               & 54.1$_{\downarrow 2.5}$  & 53.7$_{\downarrow 2.9}$  & 49.2$_{\downarrow 7.4}$  & 56.6 \\ \bottomrule
    \end{tabular}
    \caption{Results where we intend to transfer the learned language representation from CSC to news classification.}
    \label{linear-prob}
\end{table}

\subsection{Probing in Multi-Task}
To investigate the transferability from CSC to other tasks, we perform a linear probing experiment \citep{DBLP:conf/iclr/AghajanyanSGGZG21}. First, we fine-tune the model on CSC data (ECSpell). Second, we freeze the parameters of the encoder and initialize a new linear classifier following it. We fine-tune this classifer only on another task (TNEWS).
The results of the linear probing reflect whether the learned features within the encoder are generalized to transfer to new tasks.

From Table \ref{linear-prob}, we find that both tagging models suffer from a severe drop when transferring their learned language representation from CSC to TNEWS. It suggests that sequence tagging does not learn any generalized features from CSC, even degrades the language representation of the original PLM.
In contrast, we find that ReLM can transfer much better, suggesting that it retains generalized features within the language representation during fine-tuning.

\subsection{Mask Strategy}
We investigate two mask strategies of the auxiliary MLM when training ReLM. The first is to uniformly mask any characters in the sentence and the second is to mask non-error characters only.
From Table \ref{mask-strategy}, we can see that both mask strategies are effective, while masking non-error characters works better.
This is because masking error characters can reduce the amount of the real errors within the training data, which the model needs for learning correction. Additionally, we find that adding learnable prompts can further improve the performance of ReLM. The results in Table \ref{mask-strategy} and Table \ref{mask-rate} are based on ReLM with learnable prompts.

\begin{table}[t]
    \centering
    \small
    \begin{tabular}{l|cccc}
    \toprule
                       & LAW & MED & ODW & Avg \\ \midrule
        MDCSpell\textsubscript{Tagging}-MFT & 80.6&69.6 &66.9 &72.4 \\
        \midrule
        \textit{mask any} &90.5 &84.0 &82.9 &85.8 \\
        \textit{mask non error} &92.2 &85.4 &86.7 &88.1 \\ \bottomrule
    \end{tabular}
    \caption{Comparison of different mask strategies.}
    \label{mask-strategy}
\end{table}

\subsection{Mask Rate}
We also investigate the impact of the mask rate. From Table \ref{mask-rate}, we find that the performance on ECSpell keeps improving when the mask rate grows from 0\% to 30\%, and 30\% is the best choice.
While it still relies on specific data, it shows that the mask rate for ReLM is higher than that for MLM \citep{DBLP:conf/naacl/DevlinCLT19}.
This is because ReLM is essentially a further refinement of PLMs, which allows a higher mask rate.

\subsection{Case Study}
\begin{figure}[h]
    \centering
    \includegraphics[width=0.48 \textwidth]{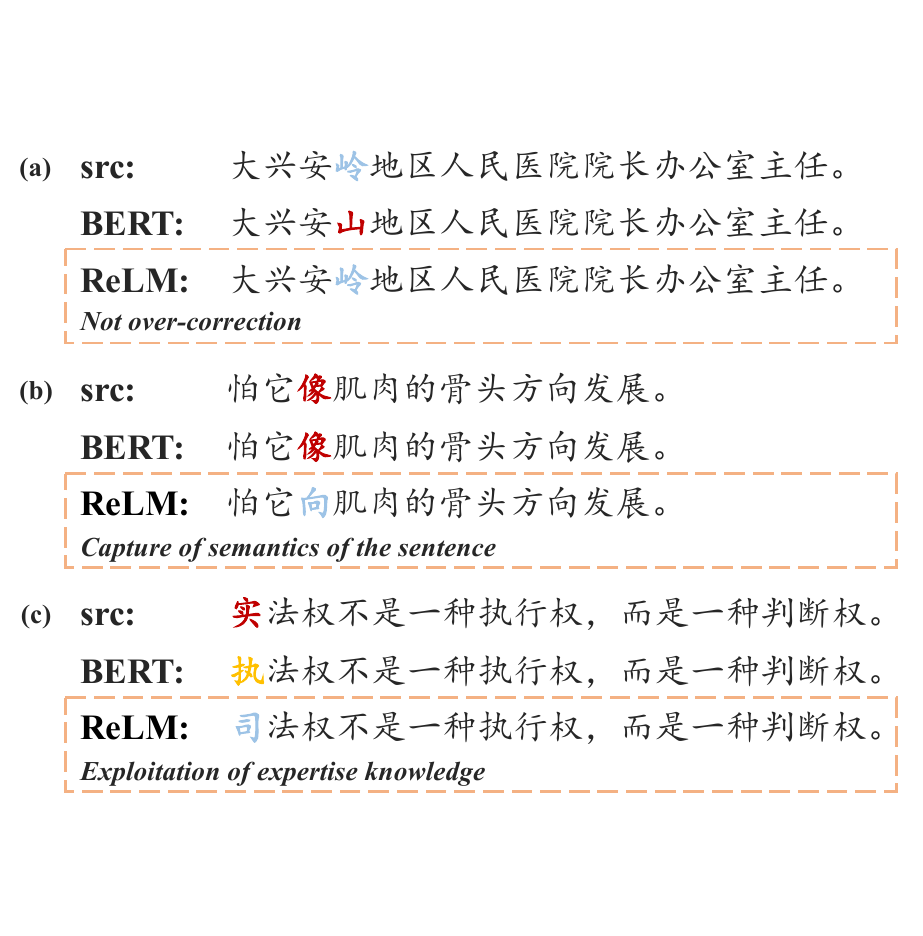}
    \caption{Cases selected from ECSpell.}
    \label{fig:case}
\end{figure}

We illustrate the superiority of ReLM over tagging through a number of cases selected from the evaluation results of masked-fine-tuned BERT-tagging and ReLM in Figure \ref{fig:case}.

\begin{table}[t]
    \centering
    \small
    \begin{tabular}{r|cccc}
    \toprule
        & LAW & MED & ODW & Avg \\ \midrule
        BERT\textsubscript{Tagging}  &37.9 &22.3 &25.0 & 28.4 \\ \midrule
        \textsc{ReLM}-0\%  &57.6 &56.9 &59.0 &57.8 \\
        10\% & 90.0 & 84.2 & 82.5 & 85.6 \\
        20\% & 91.3 & 84.8 & \textbf{86.9} & 87.7 \\ 
        30\% & \textbf{92.2} & \textbf{85.4} & 86.7 & \textbf{88.1} \\
        40\% & 91.3 & 82.8 & 84.9 & 86.3 \\
        60\% & 86.7 & 81.7 & 78.8 & 82.4 \\ \bottomrule
    \end{tabular}
    \caption{Comparison of different mask rates.}
    \label{mask-rate}
\end{table}

For the first case (\textit{Director of People's Hospital in Daxingan Ridge.}), BERT-tagging overly corrects a geopolitical place ``Daxingan Ridge'' to ``Daxingan Mountain'', which is due to the fact that it doggedly memorizes a previous edit pair ``ridge'' $\rightarrow$ ``mountain'' during training.
However, we can see that ReLM does not make this mistake.

For the second case (\textit{Worried it develops towards muscle and bone.}), it highlights the ability of ReLM to utilize the semantics of global context. Two expressions ``like'' and ``towards'' are all locally correct, while to reach the correct result, the model should refer to the word ``develop'' located at the end of the sentence (the Chinese order).

The third case is quite puzzling (\textit{Judicial power is not execution, but judgemental power.}), especially the first error. The correct answer not only necessitates the semantics but also a legal principle that ``judicial power is judgemental power'', which can only be attained through the pre-training process. We find that the tagging model does not possess such expertise and its answer is ``law enforcement power is judgemental power''.
It suggests that ReLM effectively inherits the knowledge of PLMs, while the tagging model does not even enhanced with masked-fine-tuning.

\section{Conclusion}
This papers notes a critical flaw in current CSC learning, that is conventional sequence tagging allows the correction excessively conditioned on errors, leading to limited generalizability. To address this, we propose \textit{ReLM}, where rephrasing acts as the training objective, akin to human spelling correction.
ReLM greatly outweighs previous methods on prevailing benchmarks and facilitates multi-task learning.

\section{Acknowledgements}
 This paper was partially supported by Joint Research Project of Yangtze River Delta Science and Technology Innovation Community (No. 2022CSJGG1400).

\small
\bibliography{aaai24}


\end{document}